# Ultra-Wideband Aided Fast Localization and Mapping System

Chen Wang, Handuo Zhang, Thien-Minh Nguyen, and Lihua Xie

*Abstract*—This paper proposes an ultra-wideband (UWB) aided localization and mapping system that leverages on inertial sensor and depth camera. Inspired by the fact that visual odometry (VO) system, regardless of its accuracy in the short term, still faces challenges with accumulated errors in the long run or under unfavourable environments, the UWB ranging measurements are fused to remove the visual drift and improve the robustness. A general framework is developed which consists of three parallel threads, two of which carry out the visual-inertial odometry (VIO) and UWB localization respectively. The other mapping thread integrates visual tracking constraints into a pose graph with the proposed smooth and virtual range constraints, such that an optimization is performed to provide robust trajectory estimation. Experiments show that the proposed system is able to create dense drift-free maps in real-time even running on an ultra-low power processor in featureless environments.

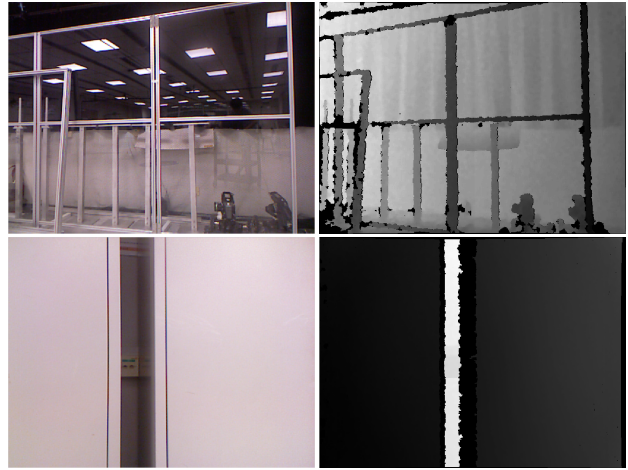

Fig. 1: The ultra-wideband aided localization and mapping system is designed to bypass the complexity of visual loop closure detection. It can be applied to the circumstance that is not suitable for a pure vision-based system. This figure shows the examples of the color (left) and depth (right) images of such challenging scenes, where large area with pseudo features and featureless regions are presented.

## I. INTRODUCTION

Real-time simultaneous localization and mapping (SLAM) system has received increasing attention in tandem with the development of smart sensors and powerful processors. As low power mobile robots such as micro unmanned aerial vehicles (MUAV) become popular, interest in equipping such low power devices with real-time localization and dense mapping system has also increased greatly.

Notable methods proposed in [1], [2] have pushed the front line forward in research on visual odometry (VO), an important research topic of SLAM. However, the VO methods such as [1], [2] leverage on matching of pixels or visual features. Hence, the map produced from these methods only consists of sparse points instead of a full depth map. Other existing methods capable of producing a real-time dense map are computationally heavy [3], [4] and are not suitable for devices with limited computational resources. In this work, we propose to employ Non-Iterative SLAM (NI-SLAM) [5] whose performance has been demonstrated to surpass that of other existing ones. NI-SLAM decreases the computational requirements while still provides accurate pose tracking by visual data association based on single key-frame training. It decouples the estimation of 6 degree-of-freedom (DoF) into several subspaces where translational and rotational motions are predicted separately and then recoupled back to the original space. This reduces the complexity significantly and enables many capabilities not yet achieved in previous works, in particular simultaneous visual tracking and dense mapping in real-time on a low power computing platform.

Having achieved a competitive SLAM method, our focus in this work shifts to the real-time dense mapping where visual loop closure is required to correct long term visual drift for NI-SLAM and in general, all vision-based SLAM methods. So far most of the works on loop closure detections are still based on visual techniques which require additional computational resources or pretrained data [6], [7]. Another work that combines VO and laser range finder was reported in [8], however it still relies on the geometry information of the environments and only reduces instead of removing visual drifts. Therefore, an alternative technique based on ultra-wideband (UWB) is investigated to bypass the complexity of visual loop closure detection. Another motivation for using UWB is to overcome the shortcomings suffered by pure visual methods. As can be seen in Fig. 1, such scenarios with reflective or featureless areas will obviously raise issues for a pure vision-based system.

As a localization technology by its own, UWB is robust to multipath and non-line-of-sight (NLOS) effects, and is able to achieve a cm-to-dm localization accuracy when fused with other sensors [9], [10]. Based on extended Kalman filter (EKF), a robot self-localization system is reported in [11] using one-way communication with fixed-position UWB modules. A tracking system proposed in [12] only requires instrumentation of the target with a single UWB transceiver.

The authors are with the School of Electrical and Electronic Engineering, Nanyang Technological University, 50 Nanyang Avenue, Singapore 639798. `wang.chen@zoho.com`; `{HZHANG032, E150040}@e.ntu.edu.sg`; `ELHXIE@ntu.edu.sg`

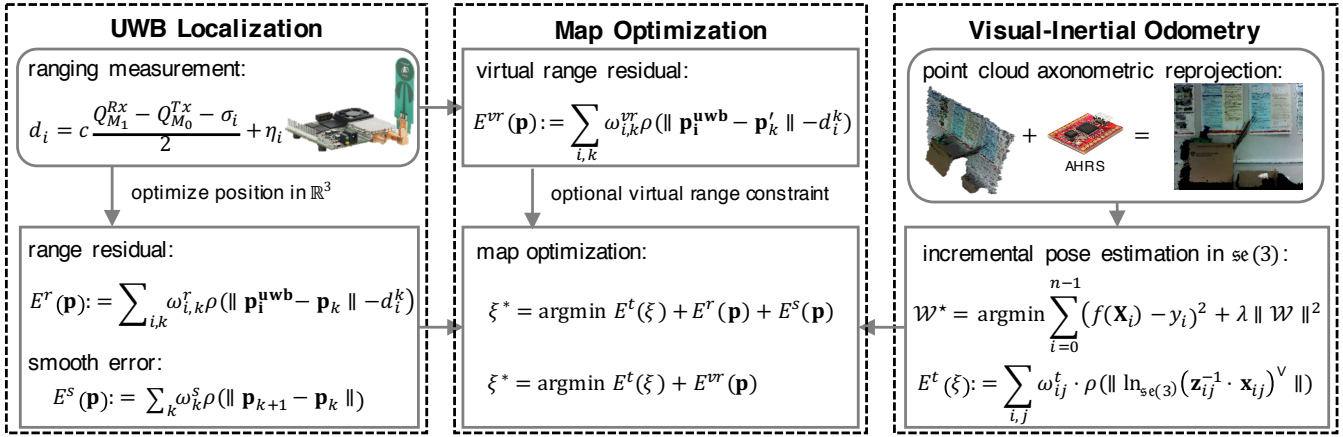

Fig. 2: **The architecture of the proposed system which consists of three parallel threads.** The UWB localization thread is to provide loop closure based on the drift-free position estimation that minimizes the range and smooth error. The visual-inertial odometry is served for local higher accurate trajectory estimation. The key frames together with the loop closure requirement is sent to the mapping thread where an optimization is performed to create live dense drift-free maps.

Note that the aforementioned works on UWB mostly rely on EKF which is sensitive to measurement outliers for highly non-linear systems. Different from the traditional filter-based methods, a graph-optimization-based framework is proposed based on a sliding window for global drift-free trajectory estimation. The optimization framework is extended further to incorporate incremental trajectory estimation from visual-inertial odometry (VIO), so that dense maps can be registered in real-time without accumulated error. To prove the generalization of the proposed framework, the use of other SLAM methods instead of NI-SLAM is also demonstrated. By comparing the performance with and without UWB aids in extreme conditions, it is shown that the proposed system is able to improve significantly the accuracy and robustness of a vision-based SLAM system, while still with feasible complexity. In summary, the main contributions are

- a graph-optimization-based framework is proposed for localization and dense reconstruction by combining the advantages of UWB and visual-inertial odometry;
- robust cost functions for UWB aided dense SLAM are designed to reduce the drift; and an axonometric map representation is came up to reduce complexity; and
- real-time dense mapping is demonstrated on an ultra-low power processor even in featureless environments.

## II. SYSTEM ARCHITECTURE

The system proposed in this paper consists of three parallel threads, namely UWB localization, VIO, and map optimization. The UWB localization thread serves to provide drift-free global position estimation. The VIO thread is to create local trajectory estimation and send the key-frames to mapping thread for optimization. Given the global drift-free coarse constraints from UWB and the incremental pose constraints from the odometry, the mapping thread directly integrates them into the pose graph where an optimization is performed to create live dense drift-free maps. The proposed system architecture is illustrated in Fig. 2, where each block will be introduced in the Section II-A, II-B, and II-C, respectively. The workflow of dense reconstruction will be presented in Section II-D.

### A. UWB Localization

*1) Cost Function:* By applying a peer-to-peer two-way time of flight measurement [9], the robot at position $\mathbf{p}_k \in \mathbb{R}^3$ is able to range to the anchors at position $\mathbf{p_i^{uwb}} \in \mathbb{R}^3$, so that distance $d_i \in \mathbb{R}$ can be obtained by the multiplication of light speed $c$ and the measurement of time of flight.

$$d_i = c\frac{Q_{M_1}^{Rx} - Q_{M_0}^{Tx} - \sigma_i}{2} + \eta_i, \quad (1a)$$
$$= \|\mathbf{p_i^{uwb}} - \mathbf{p}_k\| + \eta_i, \quad (1b)$$

where $Q_{M_1}^{Rx}$ and $Q_{M_0}^{Tx}$ is the time stamp when the UWB ranging radio is sent and received relative to the robot's clock respectively and $\sigma_i$ is the processing time delay by the anchors. It is assumed that the term $\eta_i \sim \mathcal{N}(0, \Omega_i)$ is a zero mean Gaussian noise. The proposed scheme shown in Fig. 3 is based on continuous optimization of the ranged error over a sliding window of recent poses, taking into account a constraint motion model for smoothing the estimated trajectory. The ranged error $E_r$ of a trajectory of points is defined as the sum of weighted residue $r_k^r$,

$$E_k^r(\mathbf{p}) := \sum_{i,k} \omega_{i,k}^r \cdot \rho(\underbrace{\|\mathbf{p_i^{uwb}} - \mathbf{p}_k\| - d_i^k}_{r_k^r}), \quad (2)$$

where $k$ is the time index associated with the optimizable poses $\mathbf{p}_k$ in a sliding window and $\rho(\cdot)$ is the Pseudo-Huber loss function defined as $\rho(e) = \delta^2(\sqrt{1 + (e/\delta)^2} - 1)$ that is designed to approximate quadratic function $e^2/2$ for small values and linear function $\delta e$ for large values of $e$. This ensures the derivatives are continuous for all degrees. In addition, a measurement covariance based penalty term $\omega_{i,k}^r$ is given by

$$\omega_{i,k}^r = \frac{\gamma^2}{\|\Omega_i^k\|^2 + \gamma^2}, \quad (3)$$

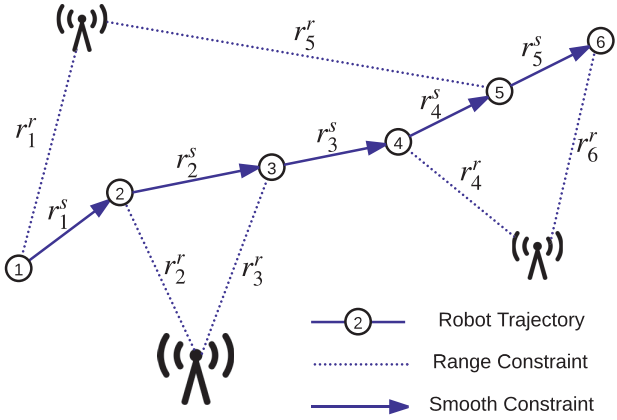

Fig. 3: **The proposed optimization scheme for UWB ranging localization.** Each time a new ranging measurement is acquired, the system updates the sliding window (by adding the new position and deleting the oldest), then optimizes the windowed trajectory based on the range and smooth constraints. Note that although a windowed trajectory is optimized in each time, only the latest pose is utilized for finding possible visual drift to obtain real-time performance.

which down-weighs the measurements with high covariance and $\gamma$ is a free-parameter. The range error (2) only gives geometric constraints on the estimated trajectory according to the range measurements, but fails to form a smooth trajectory. To overcome this problem, smooth error $E_s$ is defined among consecutive poses:

$$E_k^s(\mathbf{p}) := \sum_k \omega_k^s \cdot \rho(\underbrace{\|\mathbf{p}_{k+1} - \mathbf{p}_k\|}_{r_k^s}), \quad (4)$$

where $\rho(\cdot)$ is the Pseudo-Huber norm. Assume the maximum linear velocity of robot is $\mathbf{v}_{max} \in \mathbb{R}^3$, the penalty term $\omega_{i,k}^s$ is defined as:

$$\omega_k^s = \frac{9\gamma^2}{\|\mathbf{v}_{max}\|^2 (T_{k+1} - T_k)^2 + 9\gamma^2}, \quad (5)$$

where $T_k$ is the time stamp of pose $x_k$. This can be interpreted that the probabilistic distance between two consecutive poses is under the 3-$\sigma$ rule and the robot's velocity follows a normal distribution. Combining the range constraint (2) and smooth constraint (4), a well-defined trajectory $\mathbf{p}^*$ can be obtained through the minimization

$$\mathbf{p}^* = \arg\min_{\mathbf{p} \in \mathcal{N}_p} \underbrace{E_k^r(\mathbf{p}_k) + E_k^s(\mathbf{p}_k)}_{E_k(\mathbf{p})}, \quad (6)$$

where $\mathcal{N}_p$ is the set of positions in the sliding window.

*2) Optimization:* A numerical solution of (6) can be obtained by using the gradient descent algorithm. The residual is computed at the current stacked state $\hat{\mathbf{p}}$ by first order Taylor expansion, i.e.

$$r_k(\hat{\mathbf{p}} + \Delta \mathbf{p}) = r_k^r(\hat{\mathbf{p}} + \Delta \mathbf{p}) + r_k^s(\hat{\mathbf{p}} + \Delta \mathbf{p}), \quad (7a)$$
$$= r_k(\hat{\mathbf{p}}) + \mathbf{J}_k \Delta \mathbf{p}, \quad (7b)$$

where $\mathbf{J}_k$ is the Jacobian of $r_k(\mathbf{p})$ at $\hat{\mathbf{p}}$ and can be classified as range Jacobian $\mathbf{J}_k^r$ and and smooth Jacobian $\mathbf{J}_k^s$.

$$\mathbf{J}_k^r = \frac{\partial \|\mathbf{p}_i^{uwb} - \mathbf{p}_k\|}{\partial \mathbf{p}}, \quad \mathbf{J}_k^s = \frac{\partial \|\mathbf{p}_{k+1} - \mathbf{p}_k\|}{\partial \mathbf{p}}. \quad (8)$$

For the sake of computational efficiency, a technique used in [13] is to find a re-weighted residual. The re-weighted smooth residual is shown in (9) and a similar expression can also be found for re-weighted range residual.

$$(\omega_k^{s*} r_k^s)^T \omega_k^s (\omega_k^{s*} r_k^s) = \rho\left(\sqrt{r_k^{sT} \cdot \omega_k^s \cdot r_k^s}\right), \quad (9)$$

where

$$\omega_k^{s*} = \frac{\sqrt{\rho(\|r_k^s\|_{\omega_k^s})}}{\|r_k^s\|_{\omega_k^s}} \text{ with } \|r_k^s\|_{\omega_k^s} := \sqrt{r_k^{sT} \cdot \omega_k^s \cdot r_k^s}. \quad (10)$$

For simplicity of notation, the range and smooth errors are rewritten as combined residual $r_k^*$ relating to position $\mathbf{p}_k$. Therefore, the combined summation error $E(\mathbf{p})$ can be rewritten as,

$$E^*(\mathbf{p}) := \sum_k r_k^*(\mathbf{p}_k)^T \Omega_k^* r_k^*(\mathbf{p}_k), \quad (11)$$

where $\Omega_k^*$ is the combined weights associated with the position $\mathbf{p}_k$. Starting from a good initial value $\hat{\mathbf{p}}$, the combined range-smooth error (11) can be calculated when a small disturbance $\Delta \mathbf{p}$ is applied.

$$E_k(\hat{\mathbf{p}} + \Delta \mathbf{p}) = \sum_k r_k(\hat{\mathbf{p}} + \Delta \mathbf{p})^T \Omega_k r_k(\hat{\mathbf{p}} + \Delta \mathbf{p}). \quad (12)$$

Substitute (7b) into (12) and denote $r_k(\hat{\mathbf{p}})$ as $\hat{r}_k$:

$$= \sum_k \hat{r}_k^T \Omega_k \hat{r}_k + 2 \underbrace{\hat{r}_k^T \Omega_k \mathbf{J}_k}_{\mathbf{b}_k} \Delta \mathbf{p} + \Delta \mathbf{p}^T \underbrace{\mathbf{J}_k^T \Omega_k \mathbf{J}_k}_{\mathbf{H}_k} \Delta \mathbf{p}. \quad (13)$$

The trajectory $\mathbf{p}$ is updated by $\mathbf{p}_{(n+1)} = \mathbf{p}_{(n)} + \Delta \mathbf{p}$ in each iteration and $\Delta \mathbf{p}$ can be found via Levenberg-Marquardt algorithm [14].

$$(\mathbf{H} + \lambda \mathbf{I}) \Delta \mathbf{p}^* = -\mathbf{b}, \quad (14)$$

where $\mathbf{b} = \sum b_k$ and $\mathbf{H} = \sum \mathbf{H}_k$, $\mathbf{H} \in (\mathbb{R}^{3 \times 3})^{n \times n}$.

Each time when a new ultra-wideband radio measurement is received, the associated pose will be added to the trajectory and the two kinds of constraints will be imposed to (6). Since each pose $\mathbf{p}_k$ in the trajectory will only be restricted by one range error and two smooth errors, the $\mathbf{H}$ matrix is sparse and the number of non-zero blocks in $\mathbf{H}$ is $3n - 1$, where $n$ is the number of poses. This allows to solve (14) efficiently with sparse Cholesky factorization [15].

### B. Visual-Inertial Odometry

In this part, the previous work NI-SLAM [5] will be introduced briefly, then the estimated covariance will be derived based on some assumptions. The basic idea of NI-SLAM is that it carries out point clouds matching on the axonometric image directly. To enable this operation in real-time, the point clouds are reshaped into 1-D vectors, so that the 3 translational DoF are encoded into circular shifts of the reshaped vector, while the other 3 rotational DoF are supplemented by inertial sensors. This decoupling and reshaping technique significantly decreases the complexity compared to matching directly in the original space with 6 DoF. Let $\mathbf{X} \in \mathbb{R}^n$ be the reshaped vector of a decoupled point cloud and $\mathbf{Z} \in \mathbb{R}^n$ be its key-frame vector, where $n = M \times N$ is the axonometric image size. The matching of the two vectors is achieved by training a correlation filter [16] that regards the key-frame $\mathbf{Z}$ and its circular shifts as the training samples. Then the test sample $\mathbf{X}$ is applied to the correlation filter. The location of the maximum value in the correlation output indicates the new position of the test sample. We skip the training process, but focus on the correlation output which is denoted as $\mathbf{F}(\mathbf{X}) \in \mathbb{R}^{M \times N}$, that will be useful for calculating the estimation covariance.

Compared to UWB localization, the VIO system is able to provide 3-D incremental pose estimation, which has three additional rotational DoF in the non-Euclidean space where optimization is difficult. A proper solution is to denote the transforms on a manifold, so that they can be easily optimized in Lie algebra space. Denote the poses in the trajectory as $\mathbf{x}_k \in \mathbf{SE}(3)$ and the transformation from $\mathbf{x}_i$ to $\mathbf{x}_j$ as $\mathbf{x}_{ij} \in \mathbf{SE}(3)$, their associated representation in Lie algebra are 6-D vectors $\xi_k, \xi_{ij} \in \mathfrak{se}(3)$, for which $\xi = \left[\mathbf{p}^T, \phi^T\right]^T$, where $\mathbf{p}$ is the translational part and $\phi$ is a minimum presentation of rotation. Define the right operator $\oplus$ which applies the incremental transformation $\xi_{ij}$ on the Manifold space, i.e. $\xi_j = \xi_i \oplus \xi_{ij}$. Hence, the exponential mapping from $\mathfrak{se}(3)$ to $\mathbf{SE}(3)$ is $\mathbf{x}_i = \exp_{\mathbf{SE}(3)} \xi_i^\wedge$. Assume the incremental pose measurement from the VIO thread is denoted as $\mathbf{z}_{ij} \in \mathbf{SE}(3)$, we have

$$\mathbf{z}_{ij} = \exp_{\mathbf{SE}(3)}(\xi_{ij} \oplus \Delta\xi)^\wedge, \quad (15)$$

where $\Delta\xi$ is a small disturbance around $\xi_{ij}$ and $\Delta\xi \sim \mathcal{N}(\mathbf{0}, \Omega_{ij})$ with $\Omega_{ij} \in \mathbb{R}^{6 \times 6}$. Since the 3-D translational movements is decoupled from the rotational ones and it is assumed that the movements are independent, then $\Omega_{ij} = \text{diag}(\sigma_x, \sigma_y, \sigma_z, \Omega_\phi)$ where $\sigma_x$, $\sigma_y$, $\sigma_z$ denote the translational covariance, respectively, $\Omega_\phi \in \mathbb{R}^{3 \times 3}$ is the rotational part and can be acquired from the inertial sensor.

Now, we will derive the measurement covariance $\sigma_x$, $\sigma_y$, $\sigma_z$ from the correlation output $\mathbf{F}(\mathbf{X})$. Intuitively, the value of each element $F_{i,j}$ in $\mathbf{F}(\mathbf{X})$ indicates the estimated confidence of the translation corresponding to the location $(i, j)$. Some examples of the correlation output is shown in Fig. 4, which indicates that the correlation output $\mathbf{F}(\mathbf{X})$ can be approximated by a 2-D Gaussian function (16) with the center on $(i^*, j^*)$:

$$\mathbf{F}(i, j) = \frac{1}{2\pi\sigma_i\sigma_j} e^{-\left(\frac{(i-i^*)^2}{2\sigma_i^2} + \frac{(j-j^*)^2}{2\sigma_j^2}\right)}. \quad (16)$$

Therefore, by normalizing the correlation output $\mathbf{F}(\mathbf{X})$, we are able to compute the estimated covariance based on the maximum value $\mathbf{F}(i^*, j^*)$:

$$\sigma_x^2 = \sigma_y^2 = \frac{\epsilon^2 \cdot \sum \mathbf{F}}{2\pi \cdot \mathbf{F}(i^*, j^*)}, \quad (17)$$

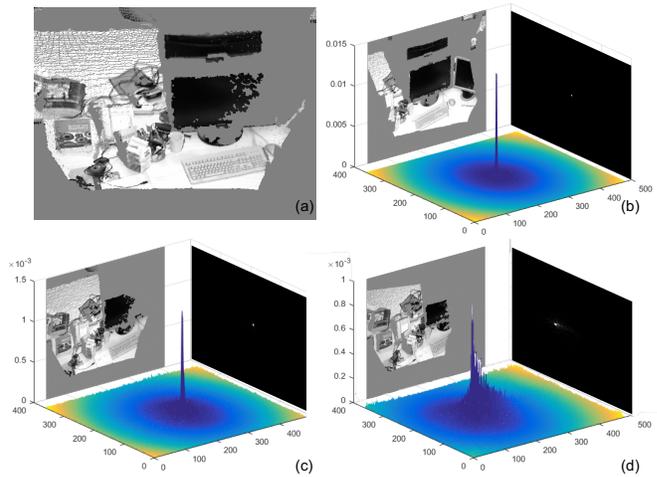

Fig. 4: **The examples of correlation output from a sequence.** (a) is the first key-frame; the top-left of figures (b), (c) and (d) are the $10_{th}$, $43_{th}$ and $188_{th}$ frames, respectively. Their correlation outputs are shown on the top-right of the corresponding figures whose PSR is 161, 71 and 41, respectively. For better visualization, the outputs are normalized and shown in 3-D space. We find that they can be approximated by a 2-D Gaussian function with the centre on the peak and this assumption is used for fast computing of estimated covariance in (17).

where $\epsilon$ is the reprojection resolution. The covariance $\sigma_z$ in depth translation can be obtained by calculating the covariance of the matched axonometric depth images directly. Let $\mathbf{X}_{\Delta x, \Delta y}$ be the shifted frame by the estimated translation, we have

$$\sigma_z^2 = \frac{Cov\left(\mathbf{Z} - \mathbf{X}_{\Delta x, \Delta y}\right)}{n}. \quad (18)$$

As the camera moves, the overlap between the key-frame and current frame will be smaller. This will result in a weak peak strength of the correlation output. In the experiments, the peak to sidelobe ratio (PSR) [17] $P_{sr} : \mathbb{R}^n \mapsto \mathbb{R}$ defined in (19) is selected to measure the peak strength.

$$P_{sr}(\mathbf{F}(\mathbf{X})) = \frac{\mathbf{F}(i^*, j^*) - \mu_s}{\sigma_s}, \quad (19)$$

where $\mu_s$ and $\sigma_s$ are respectively the mean and standard deviation of the sidelobe which is the rest of pixels excluding the peak. We found that the correlation output can be approximated by the 2-D Gaussian function especially when $P_{sr}(\mathbf{F}(\mathbf{X})) > 30$. Since a new key-frame will be created when PSR is lower than 40, the assumption can hold safely. The derived translation covariances $\sigma_x, \sigma_y, \sigma_z$ are useful for the fusion with UWB measurements presented in Section II-C. The concept of PSR will also be used for the key-frame refinement in Section II-D.

### C. UWB Aided Odometry

The minimization (6) only outputs 3-D position $\mathbf{p} \in \mathbb{R}^3$, while the visual-inertial odometry system is able to provide pose constraints in Euclidean space $\mathbf{SE}(3)$. Assume

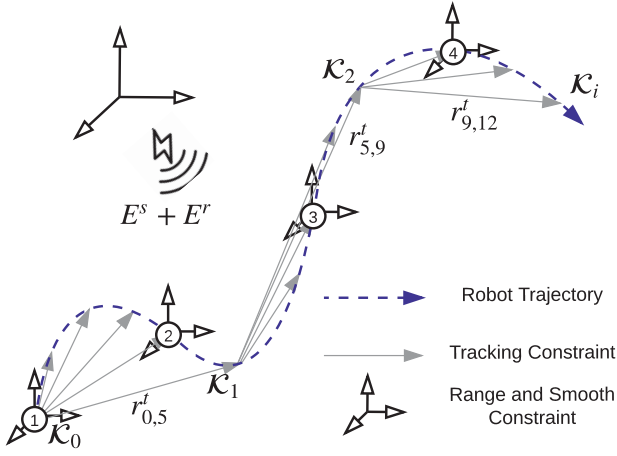
Fig. 5: **The UWB aided odometry framework based on the objective function** (22). The range-smooth constraints are applied to the nearest incremental transformation constraints $r_{i,j}^t$ relative to the key frames $\mathcal{K}_i$.

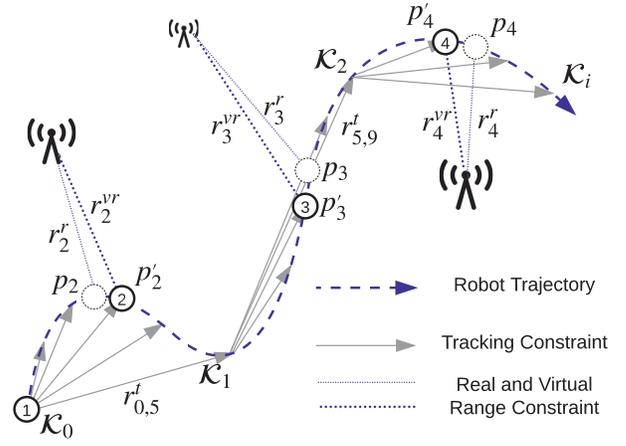
Fig. 6: **The UWB aided odometry framework based on the objective function** (25). Considering the time interval between the nearest range and tracking measurements, virtual range constraints $r_i^{vr}$ are applied to the tracking constraints.

the poses in the windowed trajectory is denoted as $\xi = [\xi_0^T, \xi_1^T, \cdots]^T$, so that the pose tracking constraint from the VIO system is defined as:

$$E^t(\xi) := \sum_{i,j} \omega_{ij}^t \cdot \rho(\underbrace{\|\ln_{\mathfrak{se}(3)}(\mathbf{z}_{ij}^{-1} \cdot \mathbf{x}_{ij})^\vee\|}_{r_{ij}^t}), \quad (20)$$

where $\vee$ is the inverse operator of $\wedge$. The weights $\omega_{ij}^t$ is defined as:

$$\omega_{ij}^t = \frac{\gamma^2}{\|\Omega_{ij}\|^2 + \gamma^2}. \quad (21)$$

The inevitable drift in (20) can be corrected by the range constraints from UWB ranging measurements. To this end, we propose to fuse the constraints based on trajectory optimization with either of the following objective functions.

*1) Cost function 1:* The most straightforward way is to apply the range constraint (2) and smooth constraint (4) together with the tracking constraints (20) directly.

$$\xi^* = \arg\min E^t(\xi) + E^r(\mathbf{p}) + E^s(\mathbf{p}). \quad (22)$$

To save computational resources, the constraints (2) and (4) are applied only when a loop closure is needed. Fig. 5 shows the basic structure of this method.

*2) Cost function 2:* Smooth error (4) provides a constraint on two consecutive poses associated with two range measurements. However, when other sensor measurements are available between the two range measurements, it is not necessary to apply this smooth constraint.

Alternatively, we apply a virtual range constraint (23) instead of the separated constraints (2) and (4), which is shown in Fig. 6. In this case, we only need to optimize the poses associated with the tracking measurements, so that more calculation can be avoided.

$$E^{vr}(\mathbf{p}) := \sum_{i,k} \omega_{i,k}^{vr} \cdot \rho(\|\mathbf{p_i^{uwb}} - \mathbf{p}_k'\| - d_i^k). \quad (23)$$

Different from (2), the virtual range error (23) applies the range constraint on position $\mathbf{p}_k'$ which is the nearest neighbour of $\mathbf{p}_k$ associated with the tracking measurement. The weights $\omega_{i,k}^{vr}$ is defined in (24), where $\Delta T_k'$ is the time interval between the nearest ranging and tracking measurements.

$$\omega_{i,k}^{vr} = \frac{9\gamma^2}{\|\mathbf{v}_{max}\|^2 (\Delta T_k')^2 + 9\|\Omega_i^k\|^2 + 9\gamma^2}. \quad (24)$$

By combining the tracking error (20) with virtual range error (23), the optimized trajectory $\xi^*$ can be obtained,

$$\xi^* = \arg\min E^t(\xi) + E^{vr}(\mathbf{p}). \quad (25)$$

*3) Optimization:* Similar to (12), the optimal solution $\xi^*$ of (22) and (25) can be found from a re-weighted residual. Starting from a initial guess $\hat{\xi}$,

$$E(\hat{\xi} \oplus \Delta\xi) = \sum_k r_k(\hat{\xi} \oplus \Delta\xi)^T \Omega_\xi r_k(\hat{\xi} \oplus \Delta\xi), \quad (26)$$

$$= E(\hat{\xi}) + J(\hat{\xi})\Delta\xi, \quad (27)$$

where $r_k$ is the re-weighted residual of (23). The trajectory $\xi$ is updated by $\xi_{(n+1)} = \xi_{(n)} \oplus \Delta\xi_{(n)}$ in each iteration and $\Delta\xi_{(n)}$ can be found by solving

$$(\mathbf{J}^T \Omega_\xi \mathbf{J} + \lambda \mathbf{I}) \Delta\xi_{(n)} = -r(\xi_{(n)})^T \Omega_\xi \mathbf{J}, \quad (28)$$

where $\mathbf{J}$ is the derivative of the stacked residual vector $\mathbf{r} = [\cdots, r_i, \cdots, r_j, \cdots]^T$ and

$$\mathbf{J} = \left.\frac{\partial \mathbf{r}(\xi_{(n)} \oplus \Delta\xi)}{\partial \Delta\xi}\right|_{\Delta\xi=\mathbf{0}}. \quad (29)$$

### D. Dense Reconstruction

The mapping thread is responsible for fusing redundant information of the point clouds, creating dense maps of the environments and visualizing the map when necessary. Section II-C shows that visual drift can be corrected by the UWB aided localization, hence no more visual loop closure is needed for dense mapping. Inspired by the fact that all the point clouds are matched with their nearest key frames on the axonometric plane, we found that maps are also able to be presented, stored and refined in the same form. Concretely, dense maps are merged by a moving average [5] with complexity $\mathcal{O}(n)$ where $n$ is the number of points to be fused. Since all the operations are pixel-wise, the operation (30) keeps much details of the map, while using less memory and computational resources. The color and depth key-frames $\mathbf{Z}^{\mathcal{C}}$ and $\mathbf{Z}^{\mathcal{D}}$ are complemented and refined by the matched frames $\mathbf{X}_k^{\mathcal{C}}$ and $\mathbf{X}_k^{\mathcal{D}}$ respectively in (30), where the operator $\mathcal{S}_k(\cdot)$ denotes the shifted image by the estimated image translation and $z$ is the estimated depth translation.

$$\mathbf{s}_k \leftarrow \mathbf{w}^{\mathbf{Z}} + \mathcal{S}_k(\mathbf{w}_k^{\mathbf{X}}) + e, \tag{30a}$$

$$\mathbf{Z}^{\mathcal{C}} \leftarrow \left(\mathbf{w}^{\mathbf{Z}} \odot \mathbf{Z}^{\mathcal{C}} + \mathcal{S}_k(\mathbf{w}_k^{\mathbf{X}} \odot \mathbf{X}_k^{\mathcal{C}})\right)/\mathbf{s}_k, \tag{30b}$$

$$\mathbf{Z}^{\mathcal{D}} \leftarrow \left(\mathbf{w}^{\mathbf{Z}} \odot \mathbf{Z}^{\mathcal{D}} + \mathcal{S}_k(\mathbf{w}_k^{\mathbf{X}} \odot (\mathbf{X}_k^{\mathcal{D}} - z))\right)/\mathbf{s}_k, \tag{30c}$$

$$\mathbf{w}^{\mathbf{z}} \leftarrow \mathbf{w}^{\mathbf{z}} + \mathcal{S}_k(\mathbf{w}_k^{\mathbf{x}}), \tag{30d}$$

where $e$ is a small scalar (set as $1e^{-7}$) to prevent division by 0. The element of the weights vector $\mathbf{w} \in \mathbb{R}^n$ presents the weight of the corresponding pixel to be fused. Each time a frame $\mathbf{X}_k$ is acquired, the corresponding weight vector $\mathbf{w}_k^{\mathbf{x}}$ is initialized by $\mathbf{w}_k^{\mathbf{x}} \leftarrow \{0, 1\}$, where 1 or 0 indicates whether the corresponding pixel can be seen in the original point cloud or not. In the experiments, this initialization is performed parallel with the axonometric reprojection. When the frame $\mathbf{X}_k$ is selected as a new key-frame, the weights vector $\mathbf{w}^{\mathbf{Z}}$ will be initialized by the weight vector of that frame, i.e. $\mathbf{w}^{\mathbf{Z}} \leftarrow \mathbf{w}_k^{\mathbf{Z}}$. Note the difference from [5] is that the key-frame refinement process (30) is only applied when $P_{sr}(\mathbf{F}(\mathbf{X})) > 100$ to ensure the fusion quality.

## III. EXPERIMENTAL RESULTS

In this section we present extensive experimental outcomes to demonstrate the superior performance of the proposed system. At first we evaluate the proposed UWB aided localization and mapping system in terms of average accuracy, dense mapping improvement, as well as the efficiency on both standard laptop and ultra-low power computing platforms. In addition, to benchmark the integration of UWB with NI-SLAM against other SLAM methods in the framework, we replace NI-SLAM with another state-of-the-art method, ORB-SLAM2 (RGB-D version) [18] and compare it with the original performance. The experimental results demonstrate the competitiveness of the proposed system as well as the versatile nature of the fusion framework.

### A. Experiment Setup

The experiment datasets were recorded by a hand-held Kinect RGB-D camera in an indoor environment. The data include color and depth images captured at 30Hz, UWB ranging measurements at 40Hz, IMU measurements at 100Hz, and 6-D ground truth obtained from a Vicon motion capture system. In order to cover distinct situations, we collected 17 datasets with different travelling distances,

TABLE I: Comparison on ATE RMSE and MAE of NI-SLAM with and without UWB aids. (Unit: m)

| Dataset | NI-SLAM+UWB | | NI-SLAM | |
|---|---|---|---|---|
| | RMSE | MAE | RMSE | MAE |
| 01_circle | **0.060** | **0.054** | 0.150 | 0.140 |
| 02_rectangle | **0.062** | **0.059** | 0.112 | 0.108 |
| 03_person1 | **0.045** | **0.035** | 0.070 | 0.065 |
| 04_person2 | **0.057** | **0.048** | 0.124 | 0.109 |
| 05_person3 | **0.054** | **0.048** | 0.120 | 0.108 |
| 06_person4 | **0.050** | **0.045** | 0.138 | 0.129 |
| 07_double_circle1 | **0.060** | **0.056** | 0.148 | 0.128 |
| 08_double_circle2 | **0.103** | **0.089** | 0.337 | 0.305 |
| 09_double_rectangle1 | **0.070** | **0.059** | 0.225 | 0.216 |
| 10_double_rectangle2 | **0.059** | **0.053** | 0.228 | 0.197 |
| 11_double_rectangle3 | **0.071** | **0.064** | 0.209 | 0.165 |
| 12_double_semicircle | **0.115** | **0.103** | 0.487 | 0.444 |
| 13_long_translation | **0.059** | **0.056** | 0.068 | 0.065 |
| 14_infinite-shape | **0.094** | **0.090** | 0.314 | 0.262 |
| 15_shake | 0.032 | 0.022 | **0.021** | **0.018** |
| 16_fast_rotation1 | **0.088** | **0.069** | 0.321 | 0.321 |
| 17_fast_rotation2 | **0.090** | **0.078** | 0.414 | 0.385 |
| mean | **0.069** | **0.060** | 0.205 | 0.186 |

speeds, dynamics, and illumination conditions in a 6m×6m area equipped with a Vicon system.

It should be highlighted that the datasets are quite challenging for pure vision-based approaches because in the scene there are large black and white regions which are featureless (Fig. 1). In addition, the several pieces of glass shown in Fig. 9 produce a lot of *pseudo landmarks* which do not follow perspective principals. This makes the traditional feature-based methods easy to lose tracking.

To test the performance on micro-robots with ultra-low power processors and compare with the state of the arts algorithms, we will use two different platforms. One is a credit-card sized mobile-phone-level UPboard® with an ultra-low power processor Atom x5-Z8350 whose scenario design power is only 2W. Running at 1.44 GHz with 2GB RAM, this platform is very difficult for most of the state-of-the-art algorithms to run in real-time. Therefore, for comparison and visualization purpose, we also test the proposed framework on a standard laptop running Ubuntu 16.04 with 8G RAM and an Intel Core i7-6700HQ CPU at 2.6 GHz. Limited by the payloads and power consumption, we choose Intel RealSense® camera to work with the lower power platform, and Microsoft Kinect® to work with the standard laptop. The IMU used in the experiments is myAHRS+®, which is a low cost high performance attitude and heading reference system (AHRS) containing 3-axis 16-bit gyroscope, a 3-axis 16-bit accelerometer, and a 3-axis 13-bit magnetometer. To obtain the best performance, NI-SLAM [5] runs with axonometric image size 360×480 for laptop and 240×360 for Up-Board®.

### B. Accuracy Evaluation

An extensive experimental validation is performed in terms of absolute trajectory error (ATE) through root mean squared error (RMSE) and median absolute error (MAE) over the entire trajectory. As shown in Table I, the proposed UWB aided system far surpasses the pure vision-based approach. Fig. 7 illustrates the plot of overhead 2-D trajectory from the No. 07 and No. 14 dataset. It can be seen that the trajectory estimation from the UWB aided system is much closer to the ground truth and no drift is presented. Because of the UWB measurement errors, it is noticed that the results without UWB aids in the No. 15 *shake* dataset performs a little better. The explanation is that the UWB constraints are better at eliminating drift in long travel rather than small tracking errors in tiny back-and-forth movements during rapid shaking. To avoid the problem, one of the solutions is to adapt the UWB aiding strategy by inserting UWB constraints only after having traveled a certain distance.

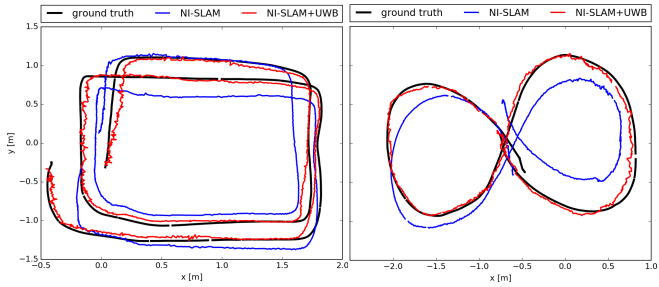

Fig. 7: **Some examples of trajectory estimation with and without UWB aids.** The trajectory estimation with UWB aids is proved to be much closer to the ground truth.

*C. Efficiency Evaluation*

We evaluate the efficiency of the proposed system on both tracking and mapping over all datasets. The average running time on two platforms is given separately in Table II. Note that the UWB process is actually independent of the SLAM thread, hence with the aid of UWB measurement, the time usage nearly does not change because of the concurrent multiple threads. Therefore, the proposed UWB aided approach does not corrupt the instantaneity of the original method. We also note that the runtime of different datasets varies according to the number of trainings. This is because NI-SLAM has to create new key-frames and train new models frequently, if the camera moves rapidly, thus increasing a little bit of running time.

*D. Dense mapping*

In this section, the performance of dense mapping is presented qualitatively. All the maps are created online and displayed without any post-processing. Note that different from the existing dense reconstruction methods [4], [19], [20], our system does not need any GPU to process large number of mapping data. In Fig. 8, the dense map of the testing area are presented. It is obvious that the point clouds align pretty well and the 3-D map depicts the environment faithfully, although there are lots of pseudo features and large featureless area in this site.

TABLE II: Average runtime of the UWB aided system. 'NI+UWB' means the UWB aided NI-SLAM system.

| Platform | Method | Tracking | Mapping | Total |
|---|---|---|---|---|
| Laptop | NI+UWB | 6.7ms | 9.1ms | 15.8ms |
| | NI-SLAM | 6.5ms | 9.2ms | 15.7ms |
| Up-Board® | NI+UWB | 27ms | 5ms | 32ms |
| | NI-SLAM | 27ms | 5ms | 32ms |

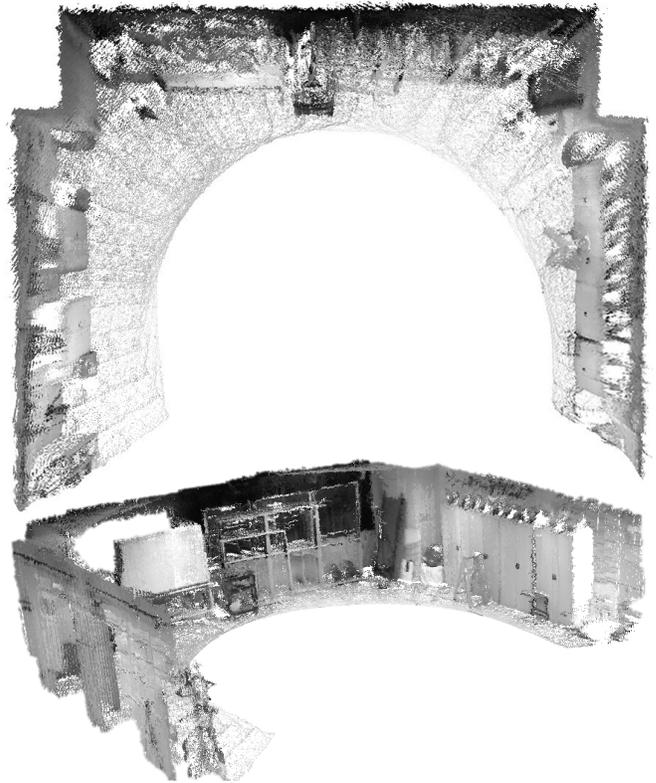

Fig. 8: **An example of dense reconstruction for the testing area.** The above two images show the dense reconstruction from two different points of view. The below image shows the panoramic view of the testing area.

*E. System Generalization*

To show the generality and robustness, performance on other SLAM methods is demonstrated in this section. One of the most popular visual SLAM methods, ORB-SLAM [18] is substituted into the proposed UWB aided system. Note that NI-SLAM is designed as a real-time visual odometry method while ORB-SLAM works as a full SLAM system with local mapping adjustment and loop-closure detection. Hence the finalized motion estimation can only be obtained after the program is terminated. To establish a low-latency system, the UWB aided system only accepts the instant pose estimation. While the loop closure detection of ORB-SLAM is still enabled to make sure the previous estimation can be corrected when a new visual loop closure is detected.

The example of feature extraction from ORB-SLAM is illustrated in Fig. 9. This shows the great challenge for ORB-SLAM since the mirror reflections taken as features actually provide highly misleading information, thus reduce localization accuracy. However, we find that the proposed framework of UWB aids successfully constraint ORB-SLAM estimates from deviating to far, according to the data in Table III. It is highlighted that NI-SLAM and ORB-SLAM are based on different front-end systems, one using on-line learning tracking and the other using ORB feature tracking. However, regardless of the difference in structure, the proposed approach can always work to provide global correction for long term errors. These experiments demonstrate the effectiveness and competitiveness of the proposed framework for localization and dense mapping.

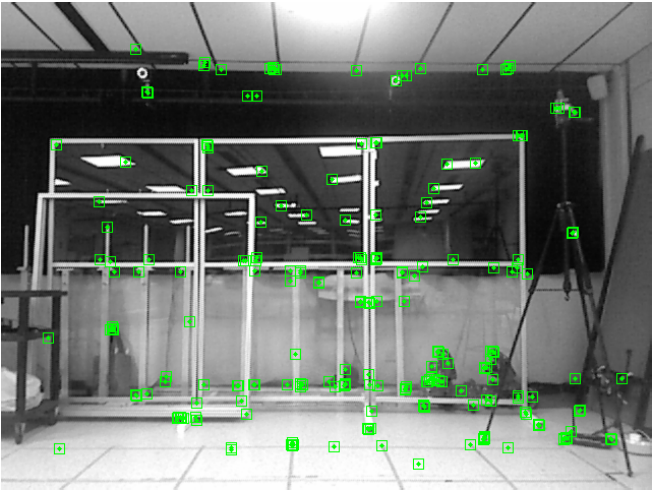

Fig. 9: **Feature extraction of landmarks by ORB-SLAM.** It is obvious to see many specular reflections in the scene which is very challenging for visual SLAM.

TABLE III: Comparison on accuracy and runtime of ORB-SLAM with and without UWB aids.

| Dataset | ORB-SLAM+UWB | | ORB-SLAM | |
|---|---|---|---|---|
| | RMSE | Runtime | RMSE | Runtime |
| 01_circle | **0.065m** | 26ms | 0.257m | 26ms |
| 02_rectangle | **0.091m** | 27ms | 0.147m | 27ms |
| 03_person1 | **0.054m** | 24ms | 0.088m | 24ms |
| 04_person2 | **0.058m** | 48ms | 0.119m | 48ms |
| 05_door | **0.112m** | 29ms | 0.432m | 29ms |
| mean | **0.076m** | 33ms | 0.209m | 33ms |

## IV. CONCLUSIONS

In this paper, an ultra-wideband aided localization and mapping system is proposed. By taking advantage of different technologies, multiple capabilities are achieved simultaneously. First, accurate and drift-free localization are enabled through a novel graph optimization technique incorporating multiple constraints from different sensors. Moreover, the proposed approach greatly reduces the complexity, so that dense maps are able to be generated in real-time even for featureless environments, while only running on an ultra-low power processor. Finally, the framework is demonstrated to be versatile, robust, and can be easily adapted for other visual SLAM methods.

ACKNOWLEDGMENT

The authors would like to thank Mr. *Junjun Wang*, Hoang Minh Chung, and Xu Fang for their help in the experiments.